# A Nonparametric Delayed Feedback Model for Conversion Rate Prediction

**Yuya Yoshikawa**[1], **Yusaku Imai**[2]
[1] STAIR Lab, Chiba Institute of Technology
[2] CyberAgent, Inc.
yoshikawa@stair.center, y24twentyfour@gmail.com

## Abstract

Predicting conversion rates (CVRs) in display advertising (e.g., predicting the proportion of users who purchase an item (i.e., a conversion) after its corresponding ad is clicked) is important when measuring the effects of ads shown to users and to understanding the interests of the users. There is generally a time delay (i.e., so-called *delayed feedback*) between the ad click and conversion. Owing to the delayed feedback, samples that are converted after an observation period may be treated as negative. To overcome this drawback, CVR prediction assuming that the time delay follows an exponential distribution has been proposed. In practice, however, there is no guarantee that the delay is generated from the exponential distribution, and the best distribution with which to represent the delay depends on the data. In this paper, we propose a nonparametric delayed feedback model for CVR prediction that represents the distribution of the time delay without assuming a parametric distribution, such as an exponential or Weibull distribution. Because the distribution of the time delay is modeled depending on the content of an ad and the features of a user, various shapes of the distribution can be represented potentially. In experiments, we show that the proposed model can capture the distribution for the time delay on a synthetic dataset, even when the distribution is complicated. Moreover, on a real dataset, we show that the proposed model outperforms the existing method that assumes an exponential distribution for the time delay in terms of conversion rate prediction.

## 1 Introduction

Predicting conversion rates (CVRs) in display advertising (e.g., the proportion of users who purchase an item (i.e., conversion) after its corresponding ad is clicked) is important when measuring the effects of advertisements shown to users and to understanding the interests of the users. Therefore, the models for CVR prediction has been well-studied for several years [Lee *et al.*, 2012; Rosales *et al.*, 2012; Ji *et al.*, 2017].

| ID | Click date | Conversion date | Ad feature 1 | ... | User feature 1 | ... |
|---|---|---|---|---|---|---|
| 1 | 2018/01/04 10:12:15 | 2018/01/12 16:30:02 | 0.1 | ... | 10.3 | ... |
| 2 | 2018/01/05 08:42:33 | unobserved | 0.3 | ... | 8.4 | ... |
| ... | ... | ... | ... | ... | ... | ... |
| n | 2018/03/30 22:50:11 | unobserved | 0.8 | ... | 12.2 | ... |

Figure 1: An example of a conversion log database. Each row (sample) includes the click date, conversion date, ad features and user features. If the sample has not yet been converted, its conversion date is recorded as 'unobserved'.

Conversion logs are stored in a database like that shown in Figure 1. Between the ad click and conversion, a time delay (i.e., the so-called *delayed feedback*) basically occurs. As a result, the conversion dates of the samples that have not been converted by the end of the observation period are recorded as 'unobserved' in the database. Because the time delay of the CVR is from hours to days, the ratio of such samples in the database is relatively large [Chappelle *et al.*, 2014].

Such samples can naively be treated as negative samples when a training conversion model is based on logistic regression or a support vector machine. However, if some of these samples are eventually converted, the model is learned on samples with incorrect labels, resulting in poor CVR prediction. It is thus necessary in practice to construct CVR prediction models considering the delayed feedback.

To deal with this drawback, Chapelle proposed a CVR prediction model that considers delayed feedback [Chapelle, 2014]. Here, the time delays of the conversion follow an exponential distribution. As shown in Figure 5 in [Chapelle, 2014], however, the actual distribution of time delays is often non-exponential and periodic, and the shape of the distribution varies by the sample. Thus, for the distribution of time delays for conversion, it is more desirable not to assume any shape of the distribution, such as an exponential or Weibull distribution, and be able to change the distribution to fit the sample.

In this paper, we propose a *nonparametric delayed feedback model* (*NoDeF* for short) for CVR prediction that represents the distribution of the time delay without assuming any parametric distributions. NoDeF partially borrows the ideas of kernel density estimation (KDE). More specifically, the distribution for the time delay is defined according to the weighted sum of kernel values of the conversion time and pseudo-points on the time axis, where the weights are learned from data. Then, as with [Chapelle, 2014], NoDeF assumes that each sample has a hidden variable indicating whether it will be converted eventually. In parameter estimation, while estimating the assignment of the hidden variable, the CVR prediction model is learned by the EM algorithm [Dempster *et al.*, 1977]. NoDeF is a general framework that can be used for not only CVR prediction in display advertisement but also various circumstances in which delayed feedback occurs.

In experiments, we demonstrate the effectiveness of the proposed model on synthetic and real datasets. On the synthetic dataset, we show that NoDeF can estimate the shape of the distribution of the time delay from data. On the real dataset, we show that NoDeF can predict the CVR for test samples, comparing with existing methods.

## 2 Related Work

As described in Section 1, NoDeF extends the delayed feedback model proposed by [Chapelle, 2014]. Their model assumes that the distribution of a time delay follows an exponential distribution with parameters determined by the input sample. [Ji *et al.*, 2017] later proposed a model for which the distribution is a mixture of Weibull distributions, and [Safari *et al.*, 2017] proposed an efficient solution for the delayed feedback model with a Weibull distribution. To the best of our knowledge, the present study is the first to represent the distribution of delayed feedback nonparametrically in the CVR prediction model.

NoDeF was inspired by an idea of [Du *et al.*, 2012] to represent time delays of information diffusion on social networks by nonparametric distributions. Because the present study deals with the problem of CVR prediction, the mathematical formulations of the previous and present papers are essentially different. Moreover, we extend the formulation of the distribution of the time delay to be able to predict the distribution for unseen samples.

Another topic well studied recently in display advertising is the multi-touch attribution setting, which considers that users are affected by the advertisement of an item via possibly multiple advertising channels, such as display, social and paid search advertising. Models that correctly measure the effect of an ad in each channel while considering the time delay between click and conversion have been proposed [Zhang *et al.*, 2015; Ji *et al.*, 2016; Ji and Wang, 2017]. Here, the distribution of the time delay in multi-touch attribution also follows an exponential or Weibull distribution. The idea of NoDeF would thus be useful for modeling multi-touch attribution.

## 3 Preliminary: Survival Analysis

This section briefly introduces the theory of survival analysis required to explain NoDeF.

Survival analysis was originally conducted to analyze the time until one or more events happen, such as the death of a biological organism or the failure of a mechanical system [Kleinbaum and Klein, 2012]. In recent years, survival analysis has also been conducted to model users' opinion formation [Yu *et al.*, 2017] and information propagation phenomena on social networks [Rodriguez *et al.*, 2013] in the area of data mining.

Table 1: Notations

| Variable | Description |
|---|---|
| $\boldsymbol{x}_i \in \mathbb{R}^M$ | A feature vector for the $i$th sample representing its ad and user, where $M$ is the length of the feature vector. |
| $y_i \in \{0,1\}$ | An observed binary value of whether the $i$th sample is converted. |
| $d_i \in [0, \infty]$ | The delay time between the click and the conversion for the $i$th sample. If not converted, $d_i = \infty$. |
| $e_i \in [0, \infty)$ | The elapsed time since the click for the $i$th sample. |
| $c_i \in \{0,1\}$ | A hidden binary value of whether the $i$th sample will be converted eventually. |
| $\boldsymbol{V} \in \mathbb{R}^{L \times M}$ | A parameter matrix for the time delay model, $L$ is the number of pseudo-points placed on time axis. |
| $\boldsymbol{w} \in \mathbb{R}^M$ | A parameter vector for the conversion model. |

Given a random variable $T > 0$ corresponding to the time that an event happens, $f(t)$ is defined as the probability density of $T$ and $F(t) = \int_0^t f(x)dx$ is its cumulative distribution function. The probability that an event does not happen up to time $t$ is given as the survival function $s(t) = 1 - F(t)$. The event rate that an event has not yet happened up to time $t$ but happens at time $t$ is then defined by a hazard function $h(t)$. The relationship among $h(t)$, $f(t)$ and $s(t)$ is

$$h(t) = \frac{f(t)}{s(t)}. \qquad (1)$$

Furthermore, the survival function $s(t)$ can be derived using hazard function $h(t)$:

$$s(t) = \exp\left(-\int_0^t h(x)dx\right). \qquad (2)$$

## 4 Proposed Model

This section explains the formulation of NoDeF and the learning algorithm for NoDeF based on the EM algorithm.

Suppose that we observe a set of $n$ quadruplets, denoted $\mathcal{D} = \{(\boldsymbol{x}_i, y_i, d_i, e_i)\}_{i=1}^n$. In particular, $\boldsymbol{x}_i \in \mathbb{R}^M$ represents the feature vector for the $i$th ad and user, where $M$ is the length of the feature vector. $y_i \in \{0,1\}$ is a binary variable indicating whether the $i$th sample is converted. $d_i \in [0, \infty]$ is the delay between the click and the conversion for the $i$th sample. If there is no conversion, $d_i = \infty$. $e_i \in [0, \infty)$ is the time elapsed since the click. Table 1 summarizes the notation of variables used for NoDeF.

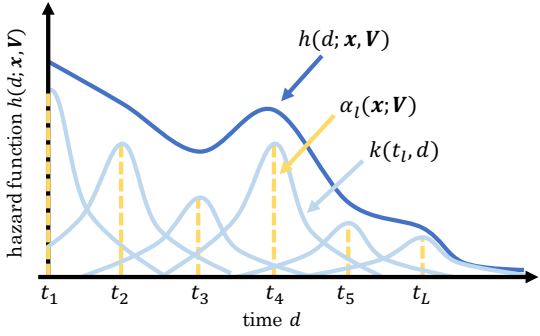

Figure 2: Relationship among the hazard function $h$, intensity function $\alpha_l$ and kernel $k$ in NoDeF. The yellow dotted line indicates $\alpha_l(\boldsymbol{x};\boldsymbol{V})$ at each time $t_l$ while the pale blue line indicates the kernel function value $k(t_l, d)$ at each time $t_l$. The dark blue line is the hazard function $h(d;\boldsymbol{x},\boldsymbol{V})$.

### 4.1 Model

NoDeF is constructed from two probabilistic models. The first model, which we refer to as the *time delay model*, is used to represent the time delays between the click and conversion. The second model, which we refer to as the *conversion model*, is a classifier that predicts whether a newly coming user is converted for a newly displayed ad. NoDeF is a joint model of these two models.

In NoDeF, we define the time delay model according to the framework of survival analysis as described in Section 3. The distribution of the time delay varies according to the contents of the ad displayed and the characteristics of the users who watch the ad. Thus, in NoDeF, we define the time delay model without assuming any distribution, such as an exponential or Weibull distribution, such that the shape of the distribution can be determined by the contents of the ad and the characteristics of the users.

We first explain the hazard function for NoDeF. To define the hazard function, NoDeF places $L \in \mathbb{N}_+$ equally spaced pseudo-points $\{t_l \geq 0\}_{l=1}^L$ on the time axis. The hazard function (i.e., the rate that the conversion for the $i$th sample has not yet happened up to time $d_i$ but happens at time $d_i$) is then defined as

$$h(d_i; \boldsymbol{x}_i, \boldsymbol{V}) = \sum_{l=1}^{L} \alpha_l(\boldsymbol{x}_i; \boldsymbol{V}) k(t_l, d_i), \quad (3)$$

where $k$ is a kernel function returning a positive value. Intuitively, the value of the kernel function represents the similarity between two points on the time axis. Here, one can use kernel functions as $k$ such that $k(t_l, \tau)$, $\int_0^a k(t_l, \tau)d\tau$ and $\int_a^\infty k(t_l, \tau)d\tau$ for $t_l, \tau, a \geq 0$ can be calculated analytically. For example, a Gaussian kernel with bandwidth parameter $h > 0$ can be used for NoDeF. In this case, the above three values are obtained as

$$k(t_l, \tau) = \exp\left(-\frac{(t_l - \tau)^2}{2h^2}\right), \quad (4)$$

$$\int_0^a k(t_l, \tau)d\tau = -h\sqrt{\frac{\pi}{2}}\left[\operatorname{erf}\left(\frac{t_l - a}{\sqrt{2}h}\right) - \operatorname{erf}\left(\frac{t_l}{\sqrt{2}h}\right)\right], (5)$$

$$\int_a^\infty k(t_l, \tau)d\tau = h\sqrt{\frac{\pi}{2}}\left[1 + \operatorname{erf}\left(\frac{t_l - a}{\sqrt{2}h}\right)\right]. \quad (6)$$

$\alpha_l$ is an intensity function with weights $\boldsymbol{V} \in \mathbb{R}^{L \times M}$, which controls a kernel value for the $l$th pseudo-point $t_l$ on the time axis, defined by

$$\alpha_l(\boldsymbol{x}_i; \boldsymbol{V}) = \left(1 + \exp\left(-\boldsymbol{V}_l^\top \boldsymbol{x}_i\right)\right)^{-1}. \quad (7)$$

Figure 2 illustrates the relationship among the hazard function $h$, intensity function $\alpha_l$ and kernel $k$ in (3).

The formulation of the hazard function (3) borrows the ideas of KDE. However, the two differ on two points. First, the hazard function controls the magnitude of the density by the intensity function $\alpha_l$ that has a feature vector $\boldsymbol{x}_i$ as input, while KDE represents the magnitude of the density by the number of data points. Second, the hazard function computes the kernel values of an input time point $d_i$ and only $L$ pseudo-points $\{t_l\}_{l=1}^L$ to calculate the density, while KDE needs to compute the kernel values of the input time point and all the observed time points whose size is generally larger than $L$. The hazard function can thus be computed quickly for a newly coming feature vector.

According to the definition of the survival function (2) and (3), the survival function for NoDeF is

$$\begin{aligned} s(d_i; \boldsymbol{x}_i, \boldsymbol{V}) &= \exp\left(-\int_0^{d_i} h(\tau; \boldsymbol{x}_i, \boldsymbol{V})d\tau\right) \quad (8) \\ &= \exp\left(-\sum_{l=1}^L \alpha_l(\boldsymbol{x}_i; \boldsymbol{V})\int_0^{d_i} k(t_l, \tau)d\tau\right). \end{aligned}$$

As in [Chapelle, 2014], we consider for each sample a hidden variable $c_i \in \{0, 1\}$, which indicates whether the $i$th sample is converted regardless of the time elapsed until the conversion. If $y_i = 1$ (i.e., the $i$th sample has been converted), then $c_i = 1$ obviously. Meanwhile, if $y_i = 0$ (i.e., the $i$th sample has not been converted before the elapsed time $e_i$), then $c_i$ cannot be determined. Thus, for samples such that $y_i = 0$, we need to estimate the assignment of $c_i$ during learning.

We then define the probability that the conversion happened at time $d_i$ for the $i$th sample. According to the hazard function (3) and the survival function (8), the probability that the conversion happened at time $d_i$ for the $i$th sample can be calculated according to

$$p(d_i|\boldsymbol{x}_i, c_i = 1) = s(d_i; \boldsymbol{x}_i, \boldsymbol{V})h(d_i; \boldsymbol{x}_i, \boldsymbol{V}). \quad (9)$$

Note that this probability (9) is undefined in the case of the sample for which $c_i = 0$.

With the conversion model, one can use any binary classifier whose likelihood is differentiable, such as logistic regression and neural network classifiers. In this paper, we simply use logistic regression defined as

$$p(c_i = 1|\boldsymbol{x}_i; \boldsymbol{w}) = \left(1 + \exp(-\boldsymbol{w}^\top \boldsymbol{x}_i)\right)^{-1}, \quad (10)$$

$$p(c_i = 0|\boldsymbol{x}_i; \boldsymbol{w}) = 1 - p(c_i = 1|\boldsymbol{x}_i), \quad (11)$$

where $\boldsymbol{w} \in \mathbb{R}^M$ is a weight vector for the conversion classifier.

We then define the likelihood for NoDeF. For the sake of convenience, we define two sets of sample indices:

$$\mathcal{I}_1 = \{i|y_i = 1, i = 1, 2, \cdots, n\}, \quad (12)$$

$$\mathcal{I}_0 = \{i|y_i = 0, i = 1, 2, \cdots, n\}. \quad (13)$$

We respectively refer to $\mathcal{I}_1$ and $\mathcal{I}_0$ as the positive sample set and negative sample set.

Given parameters $\Theta = \{V, w\}$, the likelihood of observation $\mathcal{D} = \{(x_i, y_i, d_i, e_i)\}_{i=1}^{n}$ can be factorized as

$$p(\mathcal{D}; \Theta) = \prod_{i=1}^{n} \sum_{c_i \in \{0,1\}} p(y_i|x_i, c_i, e_i) p(c_i|x_i) p(d_i|x_i, c_i = 1). \quad (14)$$

Note that if $c_i = 0$, $y_i = 0$ must be true. The equations

$$p(y_i = 0|x_i, c_i = 0, e_i) = 1, \quad (15)$$
$$p(y_i = 1|x_i, c_i = 0, e_i) = 0 \quad (16)$$

are thus satisfied consistently.

Therefore, by plugging (15) and (16) into likelihood (14) and dividing the sample indices into two sets $\mathcal{I}_1$ and $\mathcal{I}_0$, the likelihood can be deformed as

$$p(\mathcal{D}; \Theta) = \prod_{i \in \mathcal{I}_1} p(c_i = 1|x_i) p(d_i|y_i = 1, x_i) \quad (17)$$
$$\times \prod_{i \in \mathcal{I}_0} \sum_{c_i \in \{0,1\}} p(y_i = 0|x_i, c_i, e_i) p(c_i|x_i).$$

$p(y_i = 0|x_i, c_i = 1, e_i)$ is the probability that the conversion has not happened before elapsed time $e_i$ for the $i$th sample and that the conversion will happen afterward. This probability can be calculated as

$$p(y_i = 0|x_i, c_i = 1, e_i) \quad (18)$$
$$= p(d_i > e_i|x_i, c_i = 1, e_i) \quad (19)$$
$$= 1 - \int_0^{e_i} p(d_i = \tau|c_i = 1, x_i) d\tau \quad (20)$$
$$= s(e_i; x_i, V), \quad (21)$$

where the transformation from (20) to (21) is performed by applying the fact that $s(t) = 1 - F(t)$ described in Section 3.

### 4.2 Learning Algorithm

This section explains the learning algorithm for NoDeF derived on the basis of the EM algorithm.

We first define the objective function in a standard way for the EM algorithm. According to Jensen's inequality, a lower bound $Q(\Theta; \bar{\Theta})$ for the logarithm of likelihood (14) can be derived as

$$\log p(\mathcal{D}; \Theta) \geq Q(\Theta; \bar{\Theta}) \quad (22)$$
$$= \sum_{i \in \mathcal{I}_1} \log \left[ p(c_i = 1|x_i) p(d_i|x_i, c_i = 1) \right]$$
$$+ \sum_{i \in \mathcal{I}_0} \sum_{c_i \in \{0,1\}} \bar{q}_{ic_i} \log \left[ p(y_i|x_i, c_i, e_i) p(c_i|x_i) \right],$$

where $\bar{q}_{ic_i}$ is a posterior of $c_i$ defined as

$$\bar{q}_{ic} = p(c_i = c|x_i, y_i = 0, e_i) \quad (23)$$
$$\propto p(y_i = 0|x_i, c_i = c, e_i) p(c_i = c|x_i).$$

If $c = 0$, $p(y_i = 0|x_i, c_i = c, e_i)$ must have the value 1 obviously. Thus,

$$\bar{q}_{i0} = p(c_i = 0|x_i, y_i = 0, d_i, e_i) \propto p(c_i = 0|x_i), \quad (24)$$

which can be calculated using (11). Meanwhile, if $c = 1$, the posterior of $c_i$ can be calculated according to

$$\bar{q}_{i1} = p(c_i = 1|x_i, y_i = 0, d_i, e_i)$$
$$\propto p(c_i = 1|x_i) p(y_i = 0|x_i, c_i = 1, e_i), \quad (25)$$

where $p(c_i = 1|x_i)$ and $p(y_i = 0|x_i, c_i = 1, e_i)$ are respectively calculated using (10) and (18). Note that $\bar{q}_{ic}$ obtained using (24) and (25) needs to be normalized by $\bar{q}_{ic} = \bar{q}_{ic}/(\bar{q}_{i0} + \bar{q}_{i1})$. The E-step is to update the posterior of $c_i$ under the current estimates of parameters $\Theta$.

In the M-step, we update the parameters $\Theta$ employing a gradient-based optimization method such as the gradient descent method or quasi-Newton method, fixing the posterior of $c_i$. In this paper, we use L-BFGS [Liu and Nocedal, 1989], an efficient implementation of the quasi-Newton method, which needs the first-order gradients of the objective function (22) with respect to the parameters $\Theta$. We here add $\ell_2$ regularizers for $w$ and $V$ to (22), which are respectively controlled by strength parameters $\lambda_w, \lambda_V \geq 0$. These regularizers are identical to put Gaussian priors with zero mean and precision $\lambda_w$ and $\lambda_V$ for them.

First, the gradient with respect to $w$ is calculated as

$$\frac{\partial Q(\Theta; \bar{\Theta})}{\partial w} = \sum_{i \in \mathcal{I}_1} \frac{\partial}{\partial w} \log p(c_i = 1|x_i) \quad (26)$$
$$+ \sum_{i \in \mathcal{I}_0} \sum_{c_i \in \{0,1\}} \bar{q}_{ic} \frac{\partial}{\partial w} \log p(c_i|x_i) - \lambda_w w,$$

where,

$$\frac{\partial}{\partial w} \log p(c_i = 1|x_i) = x(1 - p(c_i = 1|x_i)), \quad (27)$$
$$\frac{\partial}{\partial w} \log p(c_i = 0|x_i) = -x p(c_i = 1|x_i). \quad (28)$$

Second, the gradient with respect to $V_l$ is calculated as

$$\frac{\partial Q(\Theta; \bar{\Theta})}{\partial V_l} = \sum_{i \in \mathcal{I}_1} \frac{\partial}{\partial V_l} \log p(d_i|x_i, c_i = 1) \quad (29)$$
$$+ \sum_{i \in \mathcal{I}_0} \sum_{c_i \in \{0,1\}} \bar{q}_{ic} \frac{\partial}{\partial V_l} \log p(y_i = 0|x_i, c_i, e_i) - \lambda_V V_l,$$

where,

$$\frac{\partial}{\partial V_l} \log p(d_i|x_i, c_i = 1) = \frac{\partial}{\partial V_l} \log s(d_i; x_i, V) + \frac{\partial}{\partial V_l} \log h(d_i; x_i, V), \quad (30)$$

$$\frac{\partial}{\partial V_l} \log p(y_i = 0|x_i, c_i = 1, e_i) = \frac{\partial}{\partial V_l} \log s(e_i; x_i, V), \quad (31)$$

$$\frac{\partial}{\partial V_l} \log s(d_i; x_i, V) = -x_i \alpha_l(x; V)(1 - \alpha_l(x; V)) \int_0^{d_i} k(t_l, \tau) d\tau, \quad (32)$$

$$\frac{\partial}{\partial V_l} \log h(d_i; x_i, V) = \frac{x_i \alpha_l(x; V)(1 - \alpha_l(x; V)) k(t_l, d_i)}{h(d_i; x_i, V)}. \quad (33)$$

We iteratively estimate the parameters $\Theta$ by alternating E- and M-steps until converging the objective function (22). In summary, the learning algorithm for NoDeF is shown in Algorithm 1. Here, `lbfgs_update_w` and `lbfgs_update_V` are functions that return $w$ and $V$ updated using L-BFGS, and the arguments are the initial values of the parameters and the gradients of the parameters.

**Algorithm 1** Parameter estimation of NoDeF by EM-algorithm
___
**Input:** training set $\mathcal{D}$, kernel function $k$, maximum iterations $N \in \mathbb{N}_+$, tolerance $\epsilon > 0$.
// Parameter initialization
$\boldsymbol{w}^{(0)} \sim \mathcal{N}(0,1), \ \boldsymbol{V}^{(0)} \sim \mathcal{N}(0,1), \ \Theta^{(0)} = \{\boldsymbol{w}^{(0)}, \boldsymbol{V}^{(0)}\}$
// Parameter estimation
**for** $j = 1, 2, \cdots, N$ **do**
  // E-step computation
  **for** $i \in \mathcal{I}_0$ **do**
    **for** $c \in \{0,1\}$ **do**
      $\bar{q}_{ic} = p(c_i = c | \boldsymbol{x}_i, y_i = 0, e_i)$ // using (24)
    **end for**
  **end for**
  // M-step computation
  $\boldsymbol{w}^{(j)} = \texttt{lbfgs\_update\_w}(\boldsymbol{w}^{(j-1)}, \frac{\partial}{\partial \boldsymbol{w}} Q(\Theta; \Theta^{(j-1)}))$
  $\boldsymbol{V}^{(j)} = \texttt{lbfgs\_update\_V}(\boldsymbol{V}^{(j-1)}, \frac{\partial}{\partial \boldsymbol{V}} Q(\Theta; \Theta^{(j-1)}))$
  $\Theta^{(j)} = \{\boldsymbol{w}^{(j)}, \boldsymbol{V}^{(j)}\}$
  // Checking stop condition
  **if** $Q(\Theta^{(j)}; \Theta^{(j-1)}) - Q(\Theta^{(j-1)}; \Theta^{(j-2)}) < \epsilon$ **then**
    break
  **end if**
**end for**
**Output:** $\Theta^{(j)}$
___

**Selection of hyperparameters.** NoDeF has four hyperparameters: the number of pseudo-points $L$, bandwidth for Gaussian kernel $h$, precision parameters $\lambda_{\boldsymbol{w}}$ and $\lambda_{\boldsymbol{V}}$. The hyperparameters can be determined by cross-validation or by using a validation set. As another method of determining $h$, one can use the length between two neighboring pseudo-points $t_{i+1}, t_i$ because the pseudo-points are placed at equal intervals as shown in Figure 2. We recommend using $h = (t_{i+1} - t_i)/2$ to estimate a smooth distribution for the time delay. This method is consistently used in the experiments described in Section 5.

### 4.3 Prediction

After the parameters are estimated, NoDeF can perform two types of prediction. The first is to predict whether a newly coming sample will be converted, regardless of the elapsed time of delayed feedback. In this case, the CVR can be obtained by simply applying conversion classifier (10) for the sample.

The second type of prediction is to predict whether the newly coming sample will be converted up until the time chosen. In this case, the probability that the sample $\boldsymbol{x}$ will be converted up until observation time period $E \geq 0$ can be calculated as

$$\begin{aligned} p(y=1|\boldsymbol{x}, c=1, E) &= p(c=1|\boldsymbol{x}) p(d < E | c=1, \boldsymbol{x}) \\ &= p(c=1|\boldsymbol{x}) \int_0^E p(t|c=1, \boldsymbol{x}) dt \\ &= p(c=1|\boldsymbol{x}) \left(1 - s(E; \boldsymbol{x}, \boldsymbol{V})\right) \end{aligned} \quad (34)$$

Note that the first prediction corresponds to (34) when $E = \infty$ because $\int_0^\infty p(t|c=1, \boldsymbol{x}) dt = 1$.

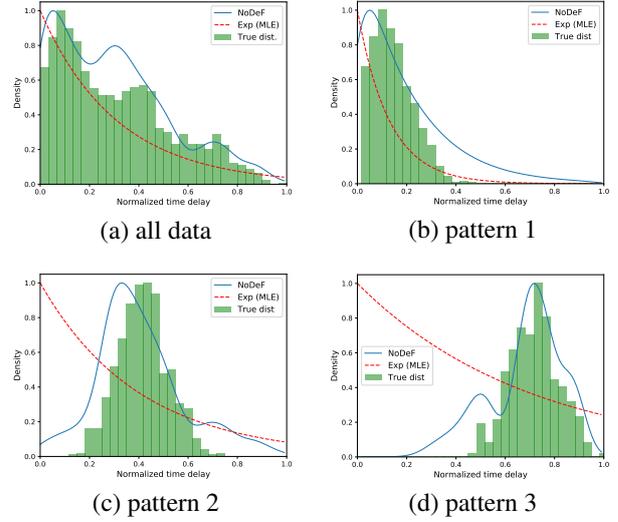

(a) all data     (b) pattern 1

(c) pattern 2     (d) pattern 3

Figure 3: Estimation results of the probability density function on the synthetic dataset. The blue solid line indicates NoDeF with $L = 40, \lambda_{\boldsymbol{w}} = 0.01, \lambda_{\boldsymbol{V}} = 0.01$, the red dashed line indicates an exponential distribution fitted by MLE, and the green histogram indicates the true distribution. The value of density for each method is normalized such that the maximum value is 1.

## 5 Experiments

This section presents the results of experiments on synthetic and real datasets.

### 5.1 Experiments on Synthetic Dataset

This subsection shows how the correct NoDeF can estimate the complicated shape of a distribution of the time delay on a synthetic dataset.

**Dataset generation.** The synthetic dataset is generated from a mixture of three different distributions so that the single dataset includes three types of conversion behavior patterns. For the first pattern, the number of samples is 100, a 10-dimensional feature vector $\boldsymbol{x} \in \mathbb{R}^{10}$ for each sample is generated from a normal distribution $\mathcal{N}(-3, 1)$, and delay time $d \geq 0$ for each sample is generated from a truncated normal distribution $\mathcal{TN}(1, 1)$ that truncates values less than 0 and greater than 10. For the second pattern, the number of samples is 70, the feature vector for each sample is generated from a normal distribution $\mathcal{N}(0, 1)$, and the delay time for each sample is generated from a truncated normal distribution $\mathcal{TN}(4, 1)$. For the third pattern, the number of samples is 30, the feature vector for each sample is generated from a normal distribution $\mathcal{N}(3, 1)$, and the delay time for each sample is generated from a truncated normal distribution $\mathcal{TN}(7, 1)$. For all samples, the elapsed time $e$ is set to 10 and label $y \in \{0, 1\}$ is determined randomly from an uniform distribution.

**Results.** Figure 3 shows the estimated density corresponding to (9) on the synthetic dataset. For comparison, we display the exponential distribution fitted by maximum likelihood estimation (MLE), which is used in [Chapelle, 2014] for modeling the time delay. As shown in Figure 3(a), NoDeF can correctly reconstruct three peaks of the true distribution. Panels (b)–(d) then show the estimated density when feature vectors for each of the three patterns are given. As shown in the figures, NoDeF can correctly estimate the density associated with each pattern using the inputted feature vectors.

Table 2: The average predictive performance for all the campaigns on the Criteo dataset. The bold face indicates the best values in each of the metrics.

|       | Log loss          | Accuracy          | AUC               |
|-------|-------------------|-------------------|-------------------|
| NAIVE | $0.3571 \pm 0.006$ | $0.8714 \pm 0.003$ | $0.7349 \pm 0.007$ |
| DFM   | $0.3450 \pm 0.008$ | $0.8702 \pm 0.004$ | **$0.7423 \pm 0.009$** |
| NoDeF | **$0.3438 \pm 0.008$** | **$0.8725 \pm 0.003$** | $0.7387 \pm 0.009$ |

Table 3: The average predictive performance for the recent campaigns on the Criteo dataset. The notation is the same as Table 2.

|       | Log loss          | Accuracy          | AUC               |
|-------|-------------------|-------------------|-------------------|
| NAIVE | $0.2818 \pm 0.021$ | $0.9124 \pm 0.013$ | $0.7187 \pm 0.019$ |
| DFM   | $0.3689 \pm 0.051$ | $0.9151 \pm 0.012$ | $0.7213 \pm 0.022$ |
| NoDeF | **$0.2575 \pm 0.020$** | **$0.9157 \pm 0.012$** | **$0.7242 \pm 0.025$** |

### 5.2 Experiments on the Criteo Dataset

This subsection presents the effectiveness of NoDeF in terms of the predictive performance on the Criteo dataset, which is a popular conversion log dataset used in [Chapelle, 2014].

**Dataset preparation.** As with [Chapelle, 2014], we use two types of the Criteo dataset. The first one includes the conversion logs associated with all the campaigns, while the second one includes those associated with the recent campaigns. For each of the two dataset, we first extract six periods. For each period, we divide the samples in the period into training, validation and test sets. The training set comprises 50,000 samples whose click date is within the first three days. Here, the conversion date is replaced with 'unobserved' when the conversion date of the samples exceeds the first three days. The validation set comprises 10,000 samples whose click date is between the final day of the training set and the next day. The test set comprises 10,000 samples whose click date is between the final day of the validation set and the next day. For the validation and test sets, the samples whose conversion date exceeds the final day of their sets are treated as negative samples.

For each sample, there are eight features taking integer values and nine categorical features. We represent each of the categorical features as a one-hot vector and then concatenate the integer features and their one-hot vectors. The dimensionality of the resulting feature vector is 2,594. We then reduce the dimensionality of the feature vectors to 100 by conducting principal component analysis.

**Setting of NoDeF.** For NoDeF, we use the normalized log values of the delay and elapsed times of each sample. The hyperparameters of NoDeF, $L \in \{10, 20, 30\}$, $\lambda_w \in \{1.0, 0.1, 0.01\}$ and $\lambda_V \in \{1.0, 0.1, 0.01\}$ are optimized using the validation set of the dataset. $h$ is determined by the method described in the last paragraph of Section 4.2.

**Setting of comparing methods.** For comparison, we use two methods that are referred to as DFM and NAIVE. These methods were used by [Chapelle, 2014]. DFM is their proposed method that considers the delayed feedback using an exponential distribution for the time delay. NAIVE is a baseline method that uses a logistic regression model [1] and treats samples for which the conversion is unobserved as negative samples. The hyperparameters in DFM and NAIVE are determined using the validation set of the dataset. For DFM, we normalize the delay time and elapse time of each sample by the observed maximum delay time in the training set.

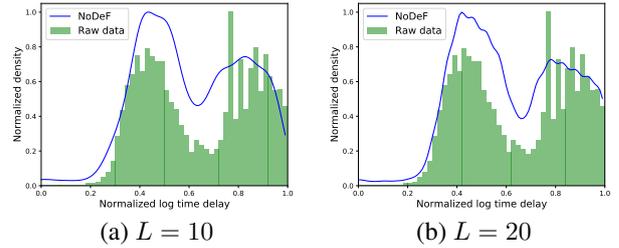

(a) $L = 10$  (b) $L = 20$

Figure 4: The estimated densities for the time delay in NoDeF with different $L$. The other hyperparameters are set to $\lambda_w = 0.1$, $\lambda_V = 0.1$. $h$ is automatically determined as described in Section 4.2.

**Results.** Tables 2 and 3 show the predictive performance on the two types of the Criteo datasets. The results indicate that NoDeF outperforms NAIVE and DFM in terms of log loss, accuracy and AUC, except for the AUC value for all the campaigns on the Criteo dataset. Then, Figure 4 shows the estimated densities for the time delay in NoDeF with different $L$. The figures shows that NoDeF could capture the two peaks appearing in the raw data without the prior knowledge of the distribution of the raw data. Then, since the densities are smooth for different $L$, this indicates that the automated determination of $h$ for NoDeF is effective.

## 6 Conclusion

We proposed a nonparametric delayed feedback model (NoDeF) for the prediction of the conversion rate in displayed advertising. Unlike existing delayed feedback models, NoDeF can estimate the distribution for the time delay between an ad click and conversion nonparametrically. By doing so, NoDeF represents complicated distributions for the time delay that cannot be captured by parametric distributions, such as exponential and Weibull distributions. In experiments, we showed that NoDeF better fitted a complicated distribution on a synthetic dataset than the existing model that assumes an exponential distribution for the time delay, and outperformed other models in terms of the predictive performance on the Criteo dataset.

In future work, to update the parameters of NoDeF as soon as possible using the latest conversion logs, we will attempt to develop an efficient learning algorithm based on the stochastic EM algorithm [Nielsen, 2000] for NoDeF. In addition, we will demonstrate the effectiveness of the idea of NoDeF by applying the model to the multi-touch attribution setting.

---

[1] We used the implementation of logistic regression in Scikit-learn 0.19.0. http://scikit-learn.org/.